\documentclass[letterpaper]{article} 
\usepackage{multirow}
\usepackage{verbatim}
\usepackage[table,xcdraw]{xcolor}
\usepackage{colortbl}
\usepackage{aaai25}  
\usepackage{times}  
\usepackage{helvet}  
\usepackage{courier}  
\usepackage[hyphens]{url}  
\usepackage{graphicx} 
\usepackage{natbib}  
\usepackage{amsmath}
\usepackage{algorithm}
\usepackage{algpseudocode}
\usepackage{arydshln}
\usepackage{graphicx}

\usepackage{caption} 
\usepackage{enumitem}

\usepackage{subcaption}
\usepackage{color}

\usepackage{newfloat}
\usepackage{listings}
\DeclareCaptionStyle{ruled}{labelfont=normalfont,labelsep=colon,strut=off} 
\floatstyle{ruled}
\newfloat{listing}{tb}{lst}{}
\floatname{listing}{Listing}
\title{Feedback-based Modal Mutual Search for Attacking \\
Vision-Language Pre-training Models}

\author {
    Renhua Ding\textsuperscript{\rm 1},
    Xinze Zhang\textsuperscript{\rm 1},
    Xiao Yang\textsuperscript{\rm 2},
    Kun He\textsuperscript{\rm 1} \thanks{\text{Corresponding author}}
}
\affiliations {
    \textsuperscript{\rm 1}School of Computer Science \& Technology, Huazhong University of Science and Technology, Wuhan, 430074, China\\
    \textsuperscript{\rm 2}Dept. of Comp. Sci. and Tech., Institute for AI, Tsinghua-Bosch Joint ML Center, THBI Lab, BNRist Center, Tsinghua University, Beijing, 100084, China.\\
    renhua@hust.edu.cn, xinze@hust.edu.cn, 
    yangxiao19@mails.tsinghua.edu.cn, brooklet60@hust.edu.cn
}

\usepackage{bibentry}

\usepackage{color}

\begin{document}

\maketitle

\begin{abstract}
Although vision-language pre-training (VLP) models have achieved remarkable progress on cross-modal tasks, they remain vulnerable to adversarial attacks. 
Using data augmentation and cross-modal interactions to generate transferable adversarial examples on surrogate models, transfer-based black-box attacks have become the mainstream methods in attacking VLP models, as they are more practical in real-world scenarios.
However, their transferability may be limited due to the differences on feature representation across different models.
To this end, we propose a new attack paradigm called Feedback-based Modal Mutual Search (FMMS).
FMMS introduces a novel modal mutual loss (MML), aiming to push away the matched image-text pairs while randomly drawing mismatched pairs closer in feature space, guiding the update directions of the adversarial examples.
Additionally, FMMS leverages the target model feedback to 
iteratively refine adversarial examples, driving them into the adversarial region.
To our knowledge, this is the first work to exploit target model feedback to explore multi-modality adversarial boundaries. 
Extensive empirical evaluations on Flickr30K and MSCOCO datasets for image-text matching tasks show that FMMS significantly outperforms the state-of-the-art baselines.
\end{abstract}
\section{Introduction}
Vision-language pre-training (VLP) models have significantly advanced cross-modal tasks by leveraging large-scale paired image-text datasets. 
These models excel in understanding visual and textual domains, achieving outstanding performance in various downstream tasks including image-text matching~\cite{image_text_matching}, image captioning~\cite{image_caption}, and visual grounding~\cite{visual_grounding}. 
However, recent researches~\cite{Sep-Attack, Vlattack, Advclip} have shown the vulnerability of VLP models to adversarial attacks, raising serious concerns about their robustness and reliability in real-world applications.
Consequently, identifying and addressing the vulnerabilities of VLP models are critical for enhancing their resilience against malicious manipulations.

\begin{figure}[t]
\centering
\includegraphics[width=3.3in]{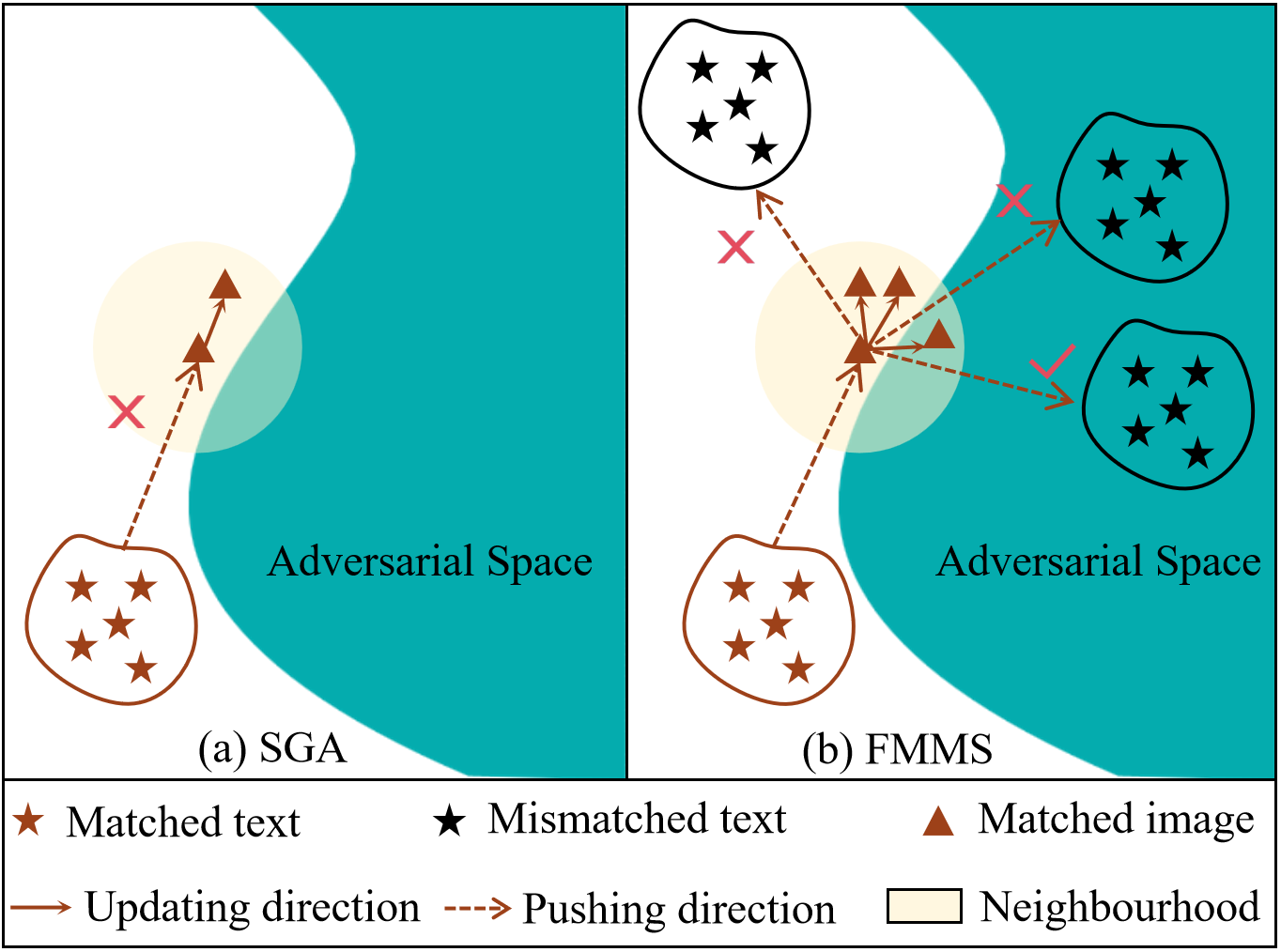}
\caption{Comparison of adversarial update directions in 
SGA and our FMMS, exemplified by updating the image modality. (a) SGA (left) only increases the distance of matched pairs, resulting in a single update direction for adversarial examples. (b) FMMS (right) additionally reduces the distance of mismatched pairs, exploring multiple update directions to effectively locate adversarial examples within the adversarial region.
}
\label{FMMS}
\end{figure}

Existing adversarial attacks can be broadly categorized into two main types: \emph{white-box attacks}~\cite{2022adversarial, 2024improving, Advclip, Vlattack} and \emph{black-box attacks}~\cite{SA-Attack, SGA, 2024boosting}. 
White-box attacks have full access to the internal structure and parameters of the target model. In contrast, black-box attacks operate with limited access to the target model, relying solely on the model's outputs to generate adversarial examples.
Black-box attacks can be further categorized into three main approaches, \emph{i.e.}, transfer-based attacks~\cite{SA-Attack, SGA, 2024boosting}, score-based attacks~\cite{2019sign, PRFA}, and decision-based attacks~\cite{QEBA, surfree}. 
Without interacting with the outputs of the target model, transfer-based attacks craft adversarial examples on a surrogate model and then 
use them to attack the target model. 
In contrast, score-based and decision-based attacks utilize the target model's confidence scores and final decisions to generate adversarial examples, respectively. 
Without requiring knowledge of the internal architecture or parameters of the target model, black-box attacks are more practical in real-world scenarios, gaining 
enormous 
attention 
in the literature. 

\begin{figure}[t]
\centering
\includegraphics[width=3.3in]{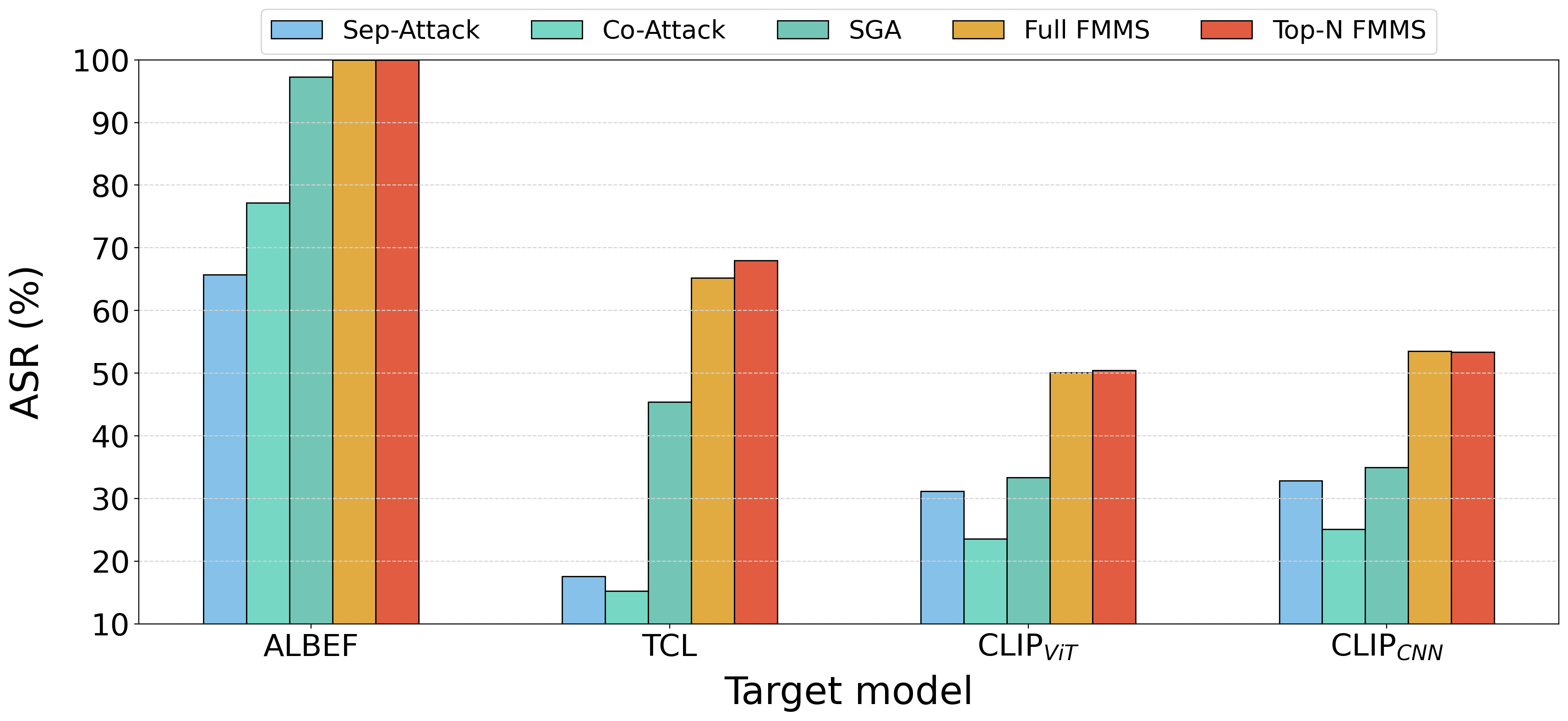}
\caption{Comparison of attack success rates (ASR) using five state-of-the-art multimodal attacks on the image-text retrieval task. Adversarial examples are generated on the surrogate model (ALBEF) to attack both white-box and black-box models. Sep-Attack combines the unimodal attack, \emph{i.e.}, PGD~\cite{PGD} and BERT-Attack~\cite{bertattack}, without cross-modal interactions. Co-Attack~\cite{Sep-Attack} only employs single-pair cross-modal interactions, while SGA utilizes data augmentation and cross-modal interactions to enhance transferability. Our Full and Top-$N$ FMMS combines cross-modal interactions and feedback information from target models to search for more efficient adversarial examples, achieving the highest ASR among various methods.
}
\label{ALBEF}
\end{figure}

Among various black-box attack approaches, transfer-based methods have been the primary focus of research efforts in attacking VLP models, owing to their practicality and simplicity of implementation. 
Recent researches~\cite{SGA, 2024boosting, OT} have explored various techniques to enhance the transferability of adversarial examples, such as data augmentation~\cite{2024boosting} and cross-modal interactions
~\cite{SGA}.
Despite practicality, these methods often encounter limitations due to differences in feature representation spaces across different VLP models, particularly between fused and aligned models. 
Fused VLP models integrate text and visual features into a unified representation, while aligned VLP models separately process different modal features. 
As a result, even when trained on the same dataset, the feature distributions of various VLP models can exhibit significant variations, especially when their architectures are different.
These discrepancies in feature representation can hinder the transferability of adversarial examples, as the examples generated on a surrogate model may not be effective on the target model due to the distinct ways features be represented and processed.
 
To address the above issue, we propose a novel method called the Feedback-based Modal Mutual Search (FMMS) attack specifically designed for multimodal tasks.
FMMS introduces 
a novel loss function termed model mutual loss (MML), that simultaneously increases the distance between matched image-text pairs and decreases the distance between mismatched pairs. MML effectively explores the target model's adversarial region, as illustrated in Figure~\ref{FMMS}. 
In addition, FMMS incorporates the feedback of the target model to refine adversarial examples with the MML loss, guiding the modal mutual search to explore more potential adversarial examples. 
FMMS employs two search strategies based on the VLP model's output information, specifically the precise matches and ranked matches of one modality retrieved by another modality:  the Full and Top-$N$ search strategies. 
The Full search considers the entire dataset as the search space, while the Top-$N$ search selects entries with a match ranking list from 1 to $N$, forming a more targeted search space that typically yields superior performance by utilizing more valuable feedback information. 
Extensive experimental results on Flickr30K and MSCOCO demonstrate that FMMS significantly outperforms existing baseline methods in generating adversarial examples across VLP models, as shown in Figure~\ref{ALBEF}. 
To the best of our knowledge, FMMS is the first work that leverages target model feedback to bridge the feature representation gap between surrogate and target models, generating more effective adversarial examples on VLP models. This provides a new perspective for multimodal adversarial attacks.



We summarize our contributions as follows:
\begin{itemize}[label=\textbullet]
    \item This work makes the first attempt to explore the VLP model's adversarial boundaries by utilizing target model feedback to perform multiple rounds of cross-modal interactions.
    \item We introduce the Feedback-based Modal Mutual Search (FMMS) 
 attack method, which designs a novel modal mutual loss (MML) and leverages the target VLP model's feedback to guide MML to refine adversarial examples, enhancing their deception.
    \item Empirical evaluations 
    demonstrate that FMMS significantly outperforms existing methods in generating effective adversarial examples across various VLP models.
\end{itemize}

\section{Related Work}
To clearly illustrate the motivation of the proposed FMMS attack for attacking VLP models, the mainstream VLP architectures and adversarial attack methods on both unimodal and multimodal models are introduced briefly.
\subsection{VLP Models}
Vision-language pre-training (VLP) models have significantly advanced multimodal artificial intelligence by enabling a joint understanding of visual and linguistic information~\cite{vlpsurvey}.
Initially relying heavily on pre-trained object detectors, these models have shifted with the introduction of Vision Transformer (ViT), which enables end-to-end image encoding by transforming images into patches~\cite{accelerating, align, blip, vlmixer}. 
VLP models are broadly categorized into two architectural paradigms: fused models and aligned models. 
Fused VLP models like ALBEF~\cite{align} and TCL~\cite{Tcl} employ separate encoders for text tokens and visual features, integrating representations with a multimodal encoder to generate unified semantics. 
However, aligned VLP models, exemplified by CLIP~\cite{Clip}, use independent encoders for each modality, and focus on aligning their feature spaces to facilitate cross-modal understanding.
This study focuses on evaluating our proposed method across various widely adopted fused or aligned VLP models.

Vision-language retrieval, also known as image-text retrieval which retrieves relevant data from one modality based on a query from another modality, is the most important cross-modal task for evaluating the effectiveness of the VLP models.
There are two subtasks: image-to-text retrieval (TR), which retrieves text that matches a given image, and text-to-image retrieval (IR), which finds the corresponding image for a given text query. 
Fused VLP models such as ALBEF~\cite{align} and TCL~\cite{Tcl} perform an initial ranking by computing semantic similarity scores for all image-text pairs, followed by a multimodal encoder to refine the ranking. 
Conversely, aligned VLP models like CLIP~\cite{Clip} generate the final rank list directly using similarity scores from the unimodal embedding space, without requiring a multimodal encoder.



\subsection{Adversarial Attacks on Unimodal Models}
Depending on the degree of acquiring target model information, adversarial attacks are generally classified into white-box and black-box attacks.
For attacking textual models, the primary methods generally involve modifying text, such as substituting specific tokens, 
\emph{e.g.}, PWWS~\cite{PWWS} and 
BERT-Attack~\cite{bertattack}.
Many white-box attack methods have achieved good performance in computer vision. The Fast Gradient Sign Method (FGSM)~\cite{FGSM} is the earliest method to utilize the sign of input gradient to maximize the classification loss and generate adversarial examples in a single step. 
The Projected Gradient Descent (PGD)~\cite{PGD} is a powerful variant following the iterative version of FGSM (I-FGSM)~\cite{I-FGSM} to produce adversarial examples. 


In black-box attacks, only the outputs of the target model can be accessed by the adversary. A mainstream black-box technique is transfer-based attacks, which generate adversarial examples on a surrogate model to deceive other models. 
Various methods have been proposed to enhance 
the transferability of adversarial examples, such as data augmentation, \emph{e.g.}, DIM~\cite{DIM}, TIM~\cite{TIM}, SIM~\cite{SIM}.
Additionally, black-box attacks can also be approached through score-based and decision-based methods. 
Score-based attacks utilize confidence scores provided by the target model to generate adversarial examples, exemplified by Zoo~\cite{ZOO} and Autozoom~\cite{autozoom}.
Decision-based attacks rely solely on the final decisions or label outputs of the target model, and iteratively modify the adversarial examples until they alter the model's decision, \emph{e.g.}, 
Boundary Attack~\cite{Boundaryattack}, SignOPT~\cite{signopt}.
Although score-based and decision-based methods can exploit information from target models, they require enormous queries and have a very high computational overhead. 
Therefore, no research has attempted score-based or decision-based attacks on VLP models so far.

\subsection{Adversarial Attacks on Multimodal Models}
For adversarial attacks on VLP models, most of the existing research centers on white-box attack settings: the separate unimodal attack (Sep-Attack) combines 
PGD~\cite{PGD} and BERT-Attack~\cite{bertattack} to attack image modality and text modality, respectively; 
the Collaborative Multimodal Adversarial Attack (Co-Attack)~\cite{Sep-Attack} considers cross-modal interactions to collectively carry out attacks on the image and text modality.
The Set-level Guidance Attack (SGA)~\cite{SGA}, which first discusses the transferability of adversarial examples on VLP models, employs data augmentation and cross-modal interactions to improve transferability. 
So far, although more transfer-based attacks~\cite{OT, 2024boosting} on VLP models have been proposed, these methods often encounter limitations due to differences in feature representation spaces across various models.
To address this, we leverage feedback from the target model over only a few rounds to perform multiple rounds of cross-modal interactions, generating more deceptive adversarial examples on surrogate models.

\section{Methodology}
In this section, we present 
the proposed Feedback-based Modal Mutual Search (FMMS) attack method. 
FMMS improves the attack performance by leveraging feedback from the target model for cross-modal searching to identify more deceptive adversarial examples. 
We first introduce relevant notations and highlight our motivation, then provide details of the proposed approach.

\subsection{Notations}
Let $(v_{i}, t_{i})$ denote the $i$-th matched image-text pair sampled from a multimodal dataset. 
Let $(v_{j}, t_{i})$ and $(v_{i}, t_{k})$ denote mismatched image-text pairs. For Vision-language pretrained (VLP) models, we denote $F_{I}$ as the image encoder and $F_{T}$ as the text encoder. $F_{I}(v)$ and $F_{T}(t)$ denote the encoded representation of the image \(v\) and text \(t\), respectively.
We define $\mathcal{B}[v, \epsilon_{v}]$ and $\mathcal{B}[t, \epsilon_{t}]$ 
as the perturbation neighborhoods for optimizing adversarial images and texts, respectively. 
$\epsilon_{v}$ and $\epsilon_{t}$ respectively denote the maximal perturbation bound for the image and text, which are configured following the previous works~\cite{SGA, Sep-Attack}.

\begin{figure}[t!]
\centering
 \begin{subfigure}[t]{0.23\textwidth}
        \centering
        \includegraphics[width=\textwidth]{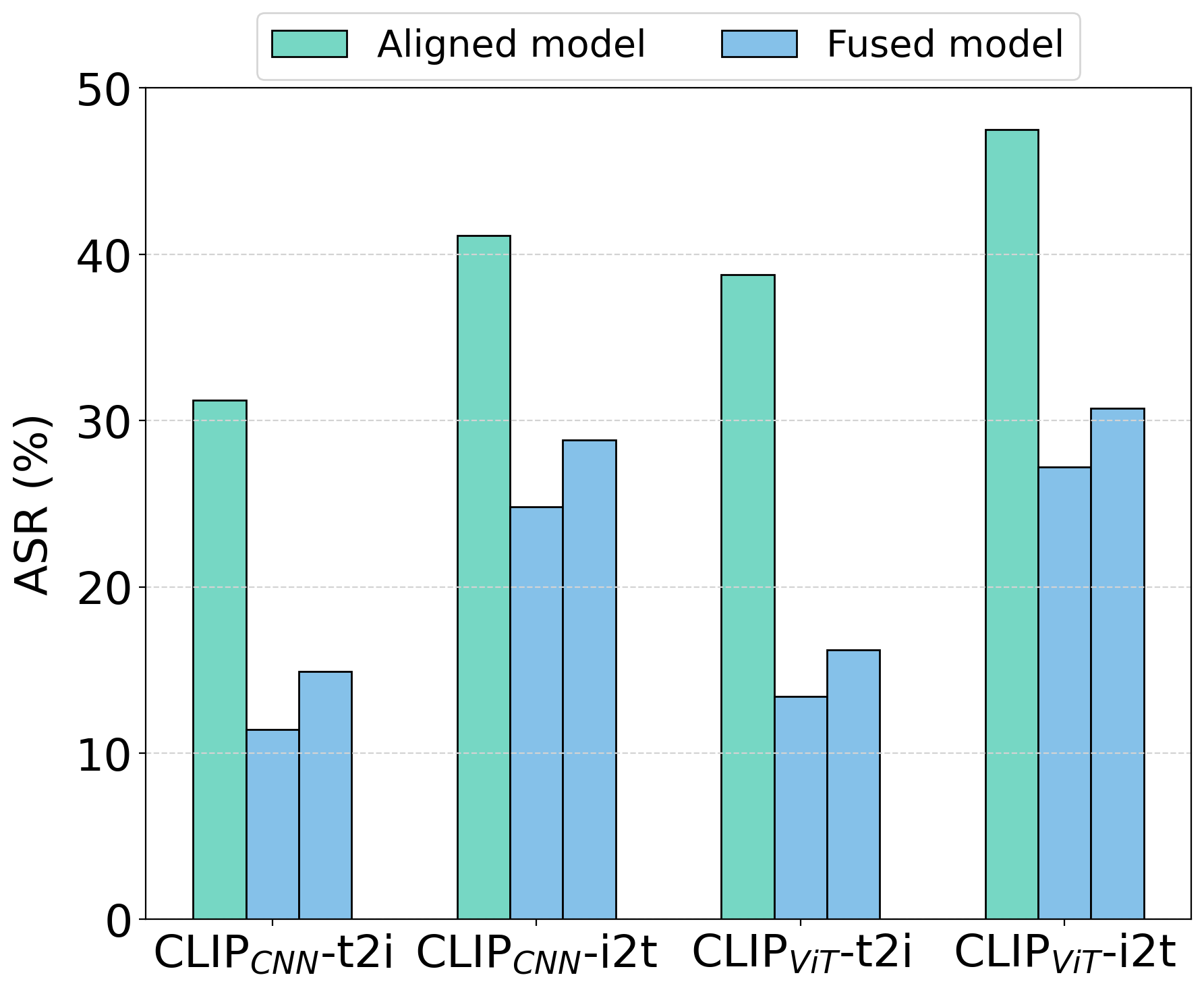}
        \caption{ }
    \end{subfigure}%
    ~ 
    \begin{subfigure}[t]{0.23\textwidth}
        \centering
        \includegraphics[width=\textwidth]{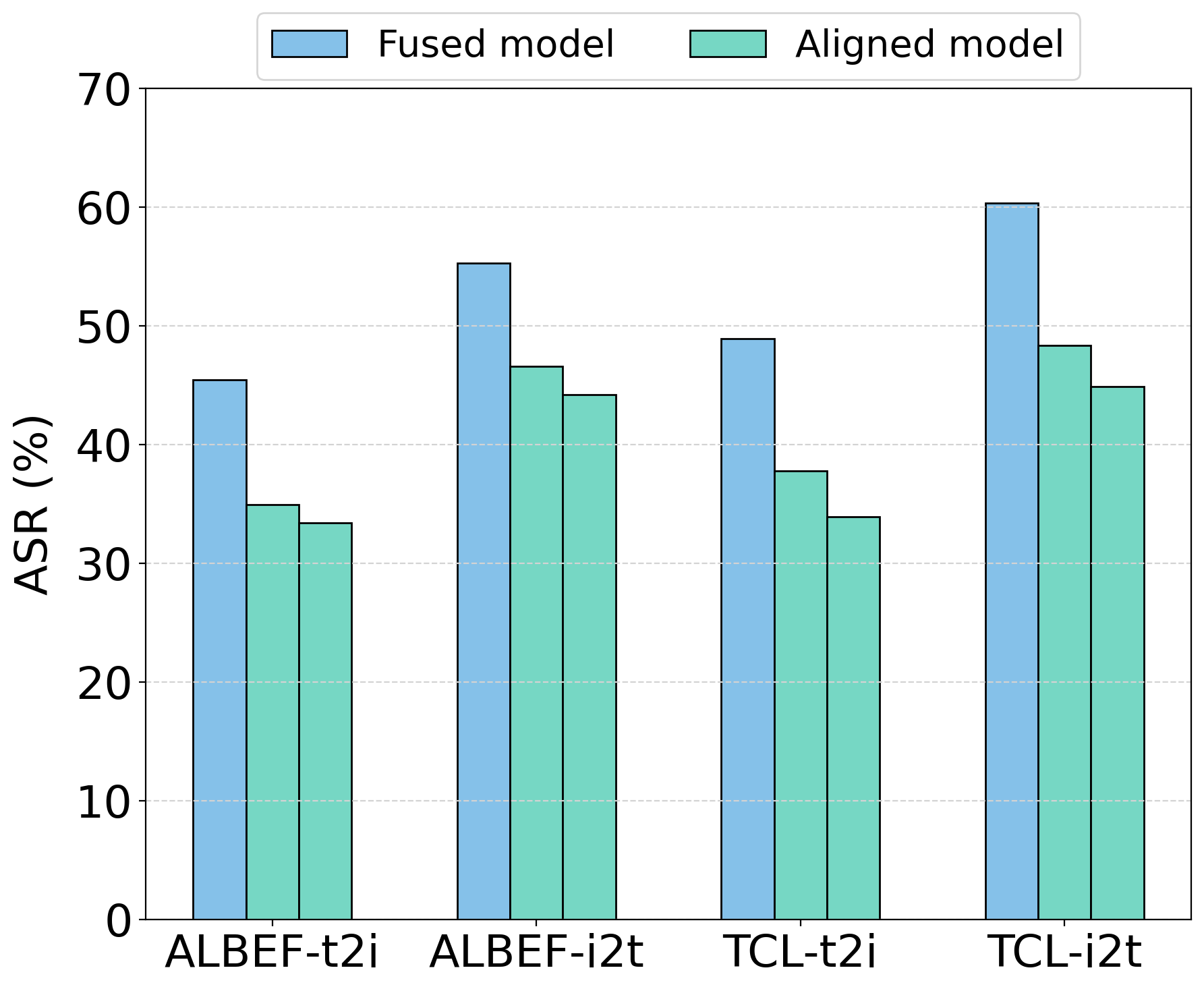}
        \caption{ }
    \end{subfigure}    
    \caption{Attack success rates (ASR) for different model architectures on image-text retrieval. Adversarial examples are crafted using four surrogate models, \emph{i.e.}, ALBEF, TCL, CLIP$_{\text{ViT}}$, and CLIP$_{\text{CNN}}$, to attack black-box fused and aligned VLP models by SGA. Different colors represent different model architectures. 
In Figure~\ref{motivation1} (a), aligned models are surrogate models and fused models are ALBEF and TCL, while in Figure~\ref{motivation1} (b), 
fused models are surrogate models and aligned models are CLIP$_{\text{CNN}}$ and CLIP$_{\text{ViT}}$, respectively.}
    \label{motivation1}
\end{figure}
\subsection{Motivation}

To investigate the transferability of multimodal adversarial examples, we first analyze the transferability of adversarial examples generated by the Set-level Guidance Attack (SGA)~\cite{SGA}.
As shown in Figure~\ref{motivation1}, we observe that when adversarial examples are generated on fused models, such as ALBEF~\cite{align} and TCL~\cite{Tcl}, they achieve higher attack success rates (ASR) against other fused models compared to the aligned models. Similarly, adversarial examples generated on aligned models, such as CLIP$_{\text{CNN}}$ and CLIP$_{\text{ViT}}$~\cite{Clip}, are more effective against other aligned models. This phenomenon indicates that differences in the feature representation space between different models, especially those with different architectures, hinder the transferability of adversarial examples.

Given this observation, a natural idea is to generate transferable adversarial examples that better match the feature representation space of the target model. 
Hence, we design a 
a modal mutual loss (MML) for leveraging the feedback of the target model to bridge the feature representation gap. 
 Besides, a multi-round cross-modal interaction is also included 
 to refine adversarial examples iteratively. In this way, we explore potential adversarial examples more effectively compared to the single-round multi-step cross-modal interaction as in SGA.

\subsection{Modal Mutual Loss}
The previous attack method of SGA~\cite{SGA} generates adversarial examples by increasing the distance between matching pairs, 
as shown in Figure~\ref{FMMS} (a). 
However, increasing the distance between matching pairs leads to a single update direction, which restricts the transferability of adversarial examples given the diverse differences among VLP models. 
Therefore, we design a modal mutual loss, defined as MML $=(\mathcal{L}_{t_{m}^{\prime}}, \mathcal{L}_{v_{adv}}, \mathcal{L}_{t_{adv}})$, to explore various update directions through multiple rounds of cross-modal interactions, generating more deceptive adversarial examples. 

Additionally, data augmentation~\cite{SGA} is another key insight identified to enhance adversarial transferability. Therefore, in our scheme, we retain alignment-preserving augmentation~\cite{SGA}, augmenting captions of each image $v_{i}$ to form the set $\boldsymbol{t_{i}}=({t_{1}, \ldots, t_{m}, \ldots, t_{M})}$ and resizing each image $v_{i}$ into different scales. We apply anti-aliasing to the scaled images~\cite{anti-aliasing}.

First, we select the mismatched image $v_{j}$ to construct  $(v_{j},t_{i})$ for each image-text pair $(v_{i},t_{i})$, generating corresponding adversarial caption set $\boldsymbol{t_{i}}^{\prime}=\left\{t_{1}^{\prime}, \ldots, t_{m}^{\prime}, \ldots, t_{M}^{\prime}\right\}$ by $\mathcal{L}_{t_{m}^{\prime}}$. 
This loss function $\mathcal{L}_{t_{m}^{\prime}}$ can be represented as:
\begin{equation}
    \begin{aligned}
        \mathcal{L}_{t_{m}^{\prime}} = 
        \frac{F_{T}\left(t_{m}\right) \cdot F_{I}(v_{j})}{\left\| F_{T}\left(t_{m}\right) \right\| \left\| F_{I}(v_{j}) \right\|}-\frac{F_{T}\left(t_{m}\right) \cdot F_{I}(v_{i})}{\left\| F_{T}\left(t_{m}\right) \right\| \left\| F_{I}(v_{i}) \right\|}
         ,
    \end{aligned}
\label{eq:1}
\end{equation}
\begin{equation}
    t_{m}^{\prime}=\underset{t_{m}^{\prime} \in B\left(t_{m}, \epsilon_{t}\right)}{\arg \max}(\mathcal{L}_{t_{m}^{\prime}}). \label{eq:tprime}
\end{equation}


Similarly, we select a mismatched caption set $\boldsymbol{t_{k}}$ to construct $(v_{i},t_{k})$, 
generating the adversarial image $v_{adv}$ by $\mathcal{L}_{v_{adv}}$, which increasing the distance of $(v_{i},t_{m}^{\prime})$ and  the distance of $(v_{i},t_{k})$ as follows:
\begin{equation}
    \begin{aligned}
        \mathcal{L}_{v_{adv}} =  &
        \left(
        \sum_{k=1}^{M} \frac{F_{T}\left(t_{k}\right)}{\left\| F_{T}\left(t_{k}\right) \right\|}
        -\sum_{m=1}^{M} \frac{F_{T}\left(t_{m}^{\prime}\right)}{\left\| F_{T}\left(t_{m}^{\prime}\right) \right\|}   
        \right) \\
        & \times \sum_{s_{i} \in S} \frac{F_{I}\left(g\left(v_{i}, s_{i}\right)\right)}{\left\| F_{I}\left(g\left(v_{i}, s_{i}\right)\right) \right\|},
    \end{aligned}
\label{eq:2}
\end{equation}
\begin{equation}
    v_{adv} = \underset{v_{adv} \in B\left[v, \epsilon_{v}\right]}{\arg \max}(\mathcal{L}_{v_{adv}}),
    \label{eq:vadv}
\end{equation}
where $g(v_{i}, s_{i})$ is an anti-aliasing scaling function that takes as input the image $v_{i}$ and the scaling factor $s_{i}$ and outputs a corresponding scaled image set. 
Finally, we construct $(t_{i}^{\prime},v_{adv})$ and $(t_{i}^{\prime},v_{j})$ from $t_{i}^{\prime}$ and $v_{adv}$ generated above, and generate  $t_{adv}$ by $\mathcal{L}_{t_{adv}}$ as follows: 

\begin{equation}
         \mathcal{L}_{t_{adv}} =  
        \frac{F_{T}\left(t_{i}^{\prime}\right) \cdot F_{I}(v_{j})}{\left\| F_{T}\left(t_{i}^{\prime}\right) \right\| \left\| F_{I}(v_{j}) \right\|}-\frac{F_{T}\left(t_{i}^{\prime}\right) \cdot F_{I}(v_{adv})}{\left\| F_{T}\left(t_{i}^{\prime}\right) \right\| \left\| F_{I}(v_{adv}) \right\|} 
          , 
\label{eq:3}
\end{equation}

\begin{equation}
    t_{adv} = \underset{t_{adv} \in B\left(t_{i}, \epsilon_{t}\right)}{\arg \max}(\mathcal{L}_{t_{adv}}).
    \label{eq:tadv}
\end{equation}

With above-mentioned three steps by Eq.~\eqref{eq:1}-Eq.~\eqref{eq:tadv}, the adversarial image \(v_{adv}\) and adversarial text \(t_{adv}\) are sequentially generated for a paired image-text example \((v_{i}, t_{i})\).

\subsection{Feedback-based Modal Mutual Search}
Different from previous black-box attacks on VLP models~\cite{SGA,2024boosting}, FMMS exploits the feedback information of the target model to search for the adversarial examples in the adversarial region. 
\begin{algorithm}[!htbp]
\caption{The FMMS Method}
\label{algorithm1}
\begin{algorithmic}[1] 
\Require 
Benign image-text pair ($v_{i}$, $t_{i}$); 
Surrogate model $f_{s}$ and target model $f_{t}$;
Search steps $T$;
\Ensure 
Adversarial image-text pair ($v_{adv}$, $t_{adv}$).

\State
Generating the initial 
adversarial image-text pair 
via Eq.~\eqref{eq:1}-Eq.~\eqref{eq:tadv}, as $t=0$.





\For{$t = 1 \to T-1$}:
\If{$f_{t}(v_{adv}) \neq t_{i}$ or $f_{t}(t_{adv}) \neq v_{i}$}
\State \Return ($v_{adv}$, $t_{adv}$);
\Else 
\If{ Top-N FMMS}
\State \textit{\textcolor{brown!60!orange}{\# Feedback the Top-N matched list}}
\State Construct search space $B_{tr}$ and $B_{ir}$ by Eq.~\eqref{eq:4} and Eq.~\eqref{eq:5}, where $N_{tr}$ and $N_{ir}$ are hyperparameters;
\Else 
\State \textit{\textcolor{brown!60!orange}{\# Feedback the hard label (Full FMMS)}}
\State Construct search space $B_{tr}$ and $B_{ir}$ by Eq.~\eqref{eq:4} and Eq.~\eqref{eq:5}, where $N_{tr}=len(V_{all})$, $N_{ir}=len(T_{all})$;
\EndIf
\State  Randomly select the mismatched image and text $v_{j} \in B_{tr}$, $t_{k} \in B_{ir}$;
\State Repeat Eq.~\eqref{eq:1}-Eq.\eqref{eq:tadv};
\State \Return ($v_{adv}$, $t_{adv}$);
\EndIf
\EndFor
\end{algorithmic}
\end{algorithm}
In the vision-language retrieval task, the target model takes inputs of image or text and returns the relevant Top-$N$ ranked instances. 
More restrictively, the target model simply outputs the hard label, \emph{i.e.}, whether it matches. 

To this end, FMMS consists of two search strategies: Full search and Top-$N$ search. The detailed pseudo-code for FMMS is shown in Algorithm~\ref{algorithm1}. 
Firstly, FMMS generates the initial adversarial image and text in line 1. 
Then, if the initial adversarial image and text attack fail, FMMS randomly selects $v_{j}$ and $t_{k}$ from the search space for modal mutual search. 
The Full search considers the entire dataset as the search space, while the Top-$N$ search attack selects entries with match rankings from 1 to $N$ to form search space $B_{tr}$ and $B_{ir}$ as follows:
\begin{equation}
    B_{tr} = top\_n(R_{f_{t}}(t_{adv}, V_{all}), N_{tr}),
\label{eq:4}
\end{equation}
\begin{equation}
    B_{ir} = top\_n(R_{f_{t}}(v_{adv}, T_{all}), N_{ir}),
\label{eq:5}
\end{equation}
where $R_{f_{t}}(\cdot,\cdot)$ calculates and returns the relevant ranked matching list of one modality retrieved by another modality on the target model $f_{t}$. 
 $V_{all}$ and $T_{all}$ denote the image set and text set in the dataset, respectively. $top\_n(\cdot, N)$ selects the top 1 to $N$ elements from the list.
Finally, repeat Eq. (1)- Eq.(6) and iterate $T-1$ times to generate the final ($v_{adv}$, $t_{adv}$).

\section{Experiments}
\subsection{Experimental Settings}
\subsubsection{Datasets}
In the experiments, we utilize two widely used datasets for multimodal tasks: Flickr30K~\cite{Flickr30k} and MSCOCO~\cite{MSCOCO}. Flickr30K comprises 31,783 images, each associated with five descriptive texts. MSCOCO consists of 123,287 images, each also paired with approximately five descriptive texts. For consistency, we used five descriptive texts per image for both datasets. Following the Karpathy splits~\cite{karpathy2015deep}, we evaluated experiments with 1K images from Flickr30K and 5K images from MSCOCO.

\subsubsection{Models}
We evaluated four popular vision-language pre-training (VLP) models: TCL~\cite{Tcl}, ALBEF~\cite{align}, CLIP${_\text{CNN}}$, and CLIP$_{\text{ViT}}$~\cite{Clip}. ALBEF incorporates three key components: an image encoder, a text encoder, and a multimodal encoder. The image encoder employs the ViT-B/16~\cite{Vit16} architecture, whereas the text and multimodal encoders are based on a 12-layer BERT~\cite{BERT} model, with the text encoder utilizing the first 6 layers and the multimodal encoder using the remaining 6 layers. TCL builds upon the ALBEF framework, introducing three contrastive learning losses to enhance the performance. CLIP integrates the image and text encoders, projecting images and texts into a shared feature space to align similar images and texts. CLIP offers two variants: CLIP$_{\text{ViT}}$, using the ViT-B/16 backbone, and CLIP${_\text{CNN}}$, based on the ResNet-101 architecture~\cite{ResNet101}.

\subsubsection{Baselines}
We adopt several widely recognized adversarial attack methods on VLP models as the baselines, including unimodal attack methods, PGD~\cite{PGD}, BERT-Attack~\cite{bertattack}, and multimodal attack methods, Sep-Attack~\cite{Sep-Attack}, Co-Attack~\cite{Sep-Attack}, and SGA~\cite{SGA}. To ensure consistent evaluation, each baseline method employs default hyperparameters provided in their public codes.

\subsubsection{Attack Settings}
Both Full FMMS and Top-$N$ FMMS are implemented in this work. For attacking image modality, we adopt PGD~\cite{PGD} with perturbation bound $\epsilon_{v}=2/255$, iteration steps $T_{v} = 10$, and step size $\alpha = 0.5/255$. The hyper-parameters of anti-aliasing scale images adopt the default settings in SGA~\cite{SGA}. For attacking text modality, we employ BERT-Attack~\cite{bertattack} with perturbation bound $\epsilon_{t} = 1$. 
For Top-$N$ FMMS, we set the length of the matched list $N_{ir} = 5$ and $N_{tr} = 10$. Moreover, we set the search steps $T = 10$ for both Full and Top-$N$ FMMS. 

\subsubsection{Metrics}
We utilize Attack Success Rate (ASR) to evaluate adversarial performance in both white-box and black-box settings. 
ASR indicates the percentage of successful adversarial examples generated. A higher ASR indicates better attack performance.
 
\begin{table*}[t]
\small
\setlength{\tabcolsep}{2mm}{
\begin{tabular}
{
>{\columncolor[HTML]{FFFFFF}}c 
>{\columncolor[HTML]{FFFFFF}}l 
>{\columncolor[HTML]{FFFFFF}}c 
>{\columncolor[HTML]{FFFFFF}}l 
>{\columncolor[HTML]{FFFFFF}}c 
>{\columncolor[HTML]{FFFFFF}}c 
>{\columncolor[HTML]{FFFFFF}}c 
>{\columncolor[HTML]{FFFFFF}}c 
>{\columncolor[HTML]{FFFFFF}}c 
>{\columncolor[HTML]{FFFFFF}}c 
>{\columncolor[HTML]{FFFFFF}}c 
>{\columncolor[HTML]{FFFFFF}}c }
\hline
\multicolumn{12}{c}{\cellcolor[HTML]{FFFFFF}Flickr30K}                                                                     \\ \hline
\multicolumn{4}{c|}{\cellcolor[HTML]{FFFFFF}Target Model (\textrightarrow)}                                                                       & \multicolumn{2}{c|}{\cellcolor[HTML]{FFFFFF}ALBEF}                                                                                               & \multicolumn{2}{c|}{\cellcolor[HTML]{FFFFFF}TCL}                                                                           & \multicolumn{2}{c|}{\cellcolor[HTML]{FFFFFF}CLIP$_{\text{ViT}}$}                                                           & \multicolumn{2}{c}{\cellcolor[HTML]{FFFFFF}CLIP$_{\text{CNN}}$}                \\ \hline
\multicolumn{2}{c|}{\cellcolor[HTML]{FFFFFF}Surrogate Model (\textdownarrow)}                        & \multicolumn{2}{c|}{\cellcolor[HTML]{FFFFFF}Attack Method (\textdownarrow)} & TR R@1                                                      & \multicolumn{1}{c|}{\cellcolor[HTML]{FFFFFF}IR R@1}                                & TR R@1                                & \multicolumn{1}{c|}{\cellcolor[HTML]{FFFFFF}IR R@1}                                & TR R@1                                & \multicolumn{1}{c|}{\cellcolor[HTML]{FFFFFF}IR R@1}                                & TR R@1                                & IR R@1                                 \\ \hline
\multicolumn{2}{c|}{\cellcolor[HTML]{FFFFFF}}                                      & \multicolumn{2}{c|}{\cellcolor[HTML]{FFFFFF}PGD}             & \cellcolor[HTML]{EFEFEF}52.45                               & \multicolumn{1}{c|}{\cellcolor[HTML]{EFEFEF}58.65}                                 & 3.06                                  & \multicolumn{1}{c|}{\cellcolor[HTML]{FFFFFF}6.79}                                  & 8.96                                  & \multicolumn{1}{c|}{\cellcolor[HTML]{FFFFFF}13.21}                                 & 10.34                                 & 14.65                                  \\
\multicolumn{2}{c|}{\cellcolor[HTML]{FFFFFF}}                                      & \multicolumn{2}{c|}{\cellcolor[HTML]{FFFFFF}BERT-Attack}     & \cellcolor[HTML]{EFEFEF}11.57                               & \multicolumn{1}{c|}{\cellcolor[HTML]{EFEFEF}27.46}                                 & 12.64                                 & \multicolumn{1}{c|}{\cellcolor[HTML]{FFFFFF}28.07}                                 & 29.33                                 & \multicolumn{1}{c|}{\cellcolor[HTML]{FFFFFF}43.17}                                 & 32.69                                 & 46.11                                  \\
\multicolumn{2}{c|}{\cellcolor[HTML]{FFFFFF}}                                      & \multicolumn{2}{c|}{\cellcolor[HTML]{FFFFFF}Sep-Attack}      & \cellcolor[HTML]{EFEFEF}65.69                               & \multicolumn{1}{c|}{\cellcolor[HTML]{EFEFEF}73.95}                                 & 17.60                                 & \multicolumn{1}{c|}{\cellcolor[HTML]{FFFFFF}32.95}                                 & 31.17                                 & \multicolumn{1}{c|}{\cellcolor[HTML]{FFFFFF}45.23}                                 & 32.82                                 & 45.49                                  \\
\multicolumn{2}{c|}{\cellcolor[HTML]{FFFFFF}}                                      & \multicolumn{2}{c|}{\cellcolor[HTML]{FFFFFF}Co-Attack}       & \cellcolor[HTML]{EFEFEF}77.16                               & \multicolumn{1}{c|}{\cellcolor[HTML]{EFEFEF}83.86}                                 & 15.21                                 & \multicolumn{1}{c|}{\cellcolor[HTML]{FFFFFF}29.49}                                 & 23.60                                 & \multicolumn{1}{c|}{\cellcolor[HTML]{FFFFFF}36.48}                                 & 25.12                                 & 38.89                                  \\
\multicolumn{2}{c|}{\cellcolor[HTML]{FFFFFF}}                                      & \multicolumn{2}{c|}{\cellcolor[HTML]{FFFFFF}SGA}             & \cellcolor[HTML]{EFEFEF}97.24                               & \multicolumn{1}{c|}{\cellcolor[HTML]{EFEFEF}97.28}                                 & 45.42                                 & \multicolumn{1}{c|}{\cellcolor[HTML]{FFFFFF}55.25}                                 & 33.38                                 & \multicolumn{1}{c|}{\cellcolor[HTML]{FFFFFF}44.16}                                 & 34.93                                 & 46.57                                  \\
\multicolumn{2}{c|}{\multirow{-6}{*}{\cellcolor[HTML]{FFFFFF}ALBEF}}               & \multicolumn{2}{c|}{\cellcolor[HTML]{FFFFFF}FMMS (ours)}            & \cellcolor[HTML]{EFEFEF}{\color[HTML]{333333} \textbf{100}} & \multicolumn{1}{c|}{\cellcolor[HTML]{EFEFEF}{\color[HTML]{333333} \textbf{97.85}}} & {\color[HTML]{333333} \textbf{67.97}} & \multicolumn{1}{c|}{\cellcolor[HTML]{FFFFFF}{\color[HTML]{333333} \textbf{58.88}}} & {\color[HTML]{333333} \textbf{50.43}} & \multicolumn{1}{c|}{\cellcolor[HTML]{FFFFFF}{\color[HTML]{333333} \textbf{46.59}}} & {\color[HTML]{333333} \textbf{53.38}} & {\color[HTML]{333333} \textbf{50.57}}  \\ \hline
\multicolumn{2}{c|}{\cellcolor[HTML]{FFFFFF}}                                      & \multicolumn{2}{c|}{\cellcolor[HTML]{FFFFFF}PGD}             & 6.15                                                        & \multicolumn{1}{c|}{\cellcolor[HTML]{FFFFFF}10.78}                                 & \cellcolor[HTML]{EFEFEF}77.87         & \multicolumn{1}{c|}{\cellcolor[HTML]{EFEFEF}79.48}                                 & 7.48                                  & \multicolumn{1}{c|}{\cellcolor[HTML]{FFFFFF}13.72}                                 & 10.34                                 & 15.33                                  \\
\multicolumn{2}{c|}{\cellcolor[HTML]{FFFFFF}}                                      & \multicolumn{2}{c|}{\cellcolor[HTML]{FFFFFF}BERT-Attack}     & 11.89                                                       & \multicolumn{1}{c|}{\cellcolor[HTML]{FFFFFF}26.82}                                 & \cellcolor[HTML]{EFEFEF}14.54         & \multicolumn{1}{c|}{\cellcolor[HTML]{EFEFEF}29.17}                                 & 29.69                                 & \multicolumn{1}{c|}{\cellcolor[HTML]{FFFFFF}44.49}                                 & 33.46                                 & 46.07                                  \\
\multicolumn{2}{c|}{\cellcolor[HTML]{FFFFFF}}                                      & \multicolumn{2}{c|}{\cellcolor[HTML]{FFFFFF}Sep-Attack}      & 20.13                                                       & \multicolumn{1}{c|}{\cellcolor[HTML]{FFFFFF}36.48}                                 & \cellcolor[HTML]{EFEFEF}84.72         & \multicolumn{1}{c|}{\cellcolor[HTML]{EFEFEF}86.07}                                 & 31.29                                 & \multicolumn{1}{c|}{\cellcolor[HTML]{FFFFFF}44.65}                                 & 33.33                                 & 45.80                                  \\
\multicolumn{2}{c|}{\cellcolor[HTML]{FFFFFF}}                                      & \multicolumn{2}{c|}{\cellcolor[HTML]{FFFFFF}Co-Attack}       & 23.15                                                       & \multicolumn{1}{c|}{\cellcolor[HTML]{FFFFFF}40.04}                                 & \cellcolor[HTML]{EFEFEF}77.94         & \multicolumn{1}{c|}{\cellcolor[HTML]{EFEFEF}85.59}                                 & 27.85                                 & \multicolumn{1}{c|}{\cellcolor[HTML]{FFFFFF}41.19}                                 & 30.74                                 & 44.11                                  \\
\multicolumn{2}{c|}{\cellcolor[HTML]{FFFFFF}}                                      & \multicolumn{2}{c|}{\cellcolor[HTML]{FFFFFF}SGA}             & 48.91                                                       & \multicolumn{1}{c|}{\cellcolor[HTML]{FFFFFF}60.34}                                 & \cellcolor[HTML]{EFEFEF}98.37         & \multicolumn{1}{c|}{\cellcolor[HTML]{EFEFEF}98.81}                                 & 33.87                                 & \multicolumn{1}{c|}{\cellcolor[HTML]{FFFFFF}44.88}                                 & 37.74                                 & 48.30                                  \\
\multicolumn{2}{c|}{\multirow{-6}{*}{\cellcolor[HTML]{FFFFFF}TCL}}                 & \multicolumn{2}{c|}{\cellcolor[HTML]{FFFFFF}FMMS (ours)}            & \textbf{69.43}                                              & \multicolumn{1}{c|}{\cellcolor[HTML]{FFFFFF}\textbf{61.48}}                        & \cellcolor[HTML]{EFEFEF}\textbf{100}  & \multicolumn{1}{c|}{\cellcolor[HTML]{EFEFEF}\textbf{99.36}}                        & \textbf{50.67}                        & \multicolumn{1}{c|}{\cellcolor[HTML]{FFFFFF}\textbf{47.68}}                        & \textbf{57.34}                        & \textbf{51.01}                         \\ \hline
\multicolumn{2}{l|}{\cellcolor[HTML]{FFFFFF}}                                      & \multicolumn{2}{c|}{\cellcolor[HTML]{FFFFFF}PGD}             & 2.50                                                        & \multicolumn{1}{c|}{\cellcolor[HTML]{FFFFFF}4.93}                                  & 4.85                                  & \multicolumn{1}{c|}{\cellcolor[HTML]{FFFFFF}8.17}                                  & \cellcolor[HTML]{EFEFEF}70.92         & \multicolumn{1}{c|}{\cellcolor[HTML]{EFEFEF}78.61}                                 & 5.36                                  & 8.44                                   \\
\multicolumn{2}{l|}{\cellcolor[HTML]{FFFFFF}}                                      & \multicolumn{2}{c|}{\cellcolor[HTML]{FFFFFF}BERT-Attack}     & 9.59                                                        & \multicolumn{1}{c|}{\cellcolor[HTML]{FFFFFF}22.64}                                 & 11.80                                 & \multicolumn{1}{c|}{\cellcolor[HTML]{FFFFFF}25.07}                                 & \cellcolor[HTML]{EFEFEF}28.34         & \multicolumn{1}{c|}{\cellcolor[HTML]{EFEFEF}39.08}                                 & 30.40                                 & 37.43                                  \\
\multicolumn{2}{l|}{\cellcolor[HTML]{FFFFFF}}                                      & \multicolumn{2}{c|}{\cellcolor[HTML]{FFFFFF}Sep-Attack}      & 9.59                                                        & \multicolumn{1}{c|}{\cellcolor[HTML]{FFFFFF}23.25}                                 & 11.38                                 & \multicolumn{1}{c|}{\cellcolor[HTML]{FFFFFF}25.60}                                 & \cellcolor[HTML]{EFEFEF}79.75         & \multicolumn{1}{c|}{\cellcolor[HTML]{EFEFEF}86.79}                                 & 30.78                                 & 39.76                                  \\
\multicolumn{2}{l|}{\cellcolor[HTML]{FFFFFF}}                                      & \multicolumn{2}{c|}{\cellcolor[HTML]{FFFFFF}Co-Attack}       & 10.57                                                       & \multicolumn{1}{c|}{\cellcolor[HTML]{FFFFFF}24.33}                                 & 11.94                                 & \multicolumn{1}{c|}{\cellcolor[HTML]{FFFFFF}26.69}                                 & \cellcolor[HTML]{EFEFEF}93.25         & \multicolumn{1}{c|}{\cellcolor[HTML]{EFEFEF}95.86}                                 & 32.52                                 & 41.82                                  \\
\multicolumn{2}{l|}{\cellcolor[HTML]{FFFFFF}}                                      & \multicolumn{2}{c|}{\cellcolor[HTML]{FFFFFF}SGA}             & 13.40                                                       & \multicolumn{1}{c|}{\cellcolor[HTML]{FFFFFF}27.22}                                 & 16.23                                 & \multicolumn{1}{c|}{\cellcolor[HTML]{FFFFFF}\textbf{30.76}}                        & \cellcolor[HTML]{EFEFEF}99.08         & \multicolumn{1}{c|}{\cellcolor[HTML]{EFEFEF}98.94}                                 & 38.76                                 & 47.79                                  \\
\multicolumn{2}{c|}{\multirow{-6}{*}{\cellcolor[HTML]{FFFFFF}CLIP$_{\text{ViT}}$}} & \multicolumn{2}{c|}{\cellcolor[HTML]{FFFFFF}FMMS (ours)}            & \textbf{20.75}                                              & \multicolumn{1}{c|}{\cellcolor[HTML]{FFFFFF}\textbf{27.27}}                        & \textbf{22.55}                        & \multicolumn{1}{c|}{\cellcolor[HTML]{FFFFFF}29.14}                                 & \cellcolor[HTML]{EFEFEF}\textbf{100}  & \multicolumn{1}{c|}{\cellcolor[HTML]{EFEFEF}\textbf{99.07}}                        & \textbf{59.00}                        & \textbf{49.13}                         \\ \hline
\multicolumn{2}{l|}{\cellcolor[HTML]{FFFFFF}}                                      & \multicolumn{2}{c|}{\cellcolor[HTML]{FFFFFF}PGD}             & 2.09                                                        & \multicolumn{1}{c|}{\cellcolor[HTML]{FFFFFF}4.82}                                  & 4.00                                  & \multicolumn{1}{c|}{\cellcolor[HTML]{FFFFFF}7.81}                                  & 1.10                                  & \multicolumn{1}{c|}{\cellcolor[HTML]{FFFFFF}6.60}                                  & \cellcolor[HTML]{EFEFEF}86.46         & \cellcolor[HTML]{EFEFEF}92.25          \\
\multicolumn{2}{l|}{\cellcolor[HTML]{FFFFFF}}                                      & \multicolumn{2}{c|}{\cellcolor[HTML]{FFFFFF}BERT-Attack}     & 8.86                                                        & \multicolumn{1}{c|}{\cellcolor[HTML]{FFFFFF}23.27}                                 & 12.33                                 & \multicolumn{1}{c|}{\cellcolor[HTML]{FFFFFF}25.48}                                 & 27.12                                 & \multicolumn{1}{c|}{\cellcolor[HTML]{FFFFFF}37.44}                                 & \cellcolor[HTML]{EFEFEF}30.40         & \cellcolor[HTML]{EFEFEF}40.10          \\
\multicolumn{2}{l|}{\cellcolor[HTML]{FFFFFF}}                                      & \multicolumn{2}{c|}{\cellcolor[HTML]{FFFFFF}Sep-Attack}      & 8.55                                                        & \multicolumn{1}{c|}{\cellcolor[HTML]{FFFFFF}23.41}                                 & 12.64                                 & \multicolumn{1}{c|}{\cellcolor[HTML]{FFFFFF}26.12}                                 & 28.34                                 & \multicolumn{1}{c|}{\cellcolor[HTML]{FFFFFF}39.43}                                 & \cellcolor[HTML]{EFEFEF}91.44         & \cellcolor[HTML]{EFEFEF}95.44          \\
\multicolumn{2}{l|}{\cellcolor[HTML]{FFFFFF}}                                      & \multicolumn{2}{c|}{\cellcolor[HTML]{FFFFFF}Co-Attack}       & 8.79                                                        & \multicolumn{1}{c|}{\cellcolor[HTML]{FFFFFF}23.74}                                 & 13.10                                 & \multicolumn{1}{c|}{\cellcolor[HTML]{FFFFFF}26.07}                                 & 28.79                                 & \multicolumn{1}{c|}{\cellcolor[HTML]{FFFFFF}40.03}                                 & \cellcolor[HTML]{EFEFEF}94.76         & \cellcolor[HTML]{EFEFEF}96.89          \\
\multicolumn{2}{l|}{\cellcolor[HTML]{FFFFFF}}                                      & \multicolumn{2}{c|}{\cellcolor[HTML]{FFFFFF}SGA}             & 11.42                                                       & \multicolumn{1}{c|}{\cellcolor[HTML]{FFFFFF}24.80}                                 & 14.91                                 & \multicolumn{1}{c|}{\cellcolor[HTML]{FFFFFF}28.82}                                 & 31.24                                 & \multicolumn{1}{c|}{\cellcolor[HTML]{FFFFFF}42.12}                                 & \cellcolor[HTML]{EFEFEF}99.24         & \cellcolor[HTML]{EFEFEF}\textbf{99.49} \\
\multicolumn{2}{c|}{\multirow{-6}{*}{\cellcolor[HTML]{FFFFFF}CLIP$_{\text{CNN}}$}} & \multicolumn{2}{c|}{\cellcolor[HTML]{FFFFFF}FMMS (ours)}            & \textbf{17.73}                                              & \multicolumn{1}{c|}{\cellcolor[HTML]{FFFFFF}\textbf{25.77}}                        & \textbf{19.60}                         & \multicolumn{1}{c|}{\cellcolor[HTML]{FFFFFF}\textbf{27.88}}                        & \textbf{47.48}                        & \multicolumn{1}{c|}{\cellcolor[HTML]{FFFFFF}\textbf{44.10}}                        & \cellcolor[HTML]{EFEFEF}\textbf{100}  & \cellcolor[HTML]{EFEFEF}99.36          \\ \hline
\end{tabular}%
}

\caption{Comparison with existing attack methods on the image-text retrieval task on Flickr30K. This table presents the attack success rate (ASR, \%) R@1 for TR and IR. The gray background cells in the table indicate white-box attacks. A higher ASR in black-box attacks denotes better adversarial performance. The best results appear in \textbf{bold}.}
\label{Table2}
\end{table*}

\subsection{Experimental Results}
\subsubsection{Comparison with the Baselines} 

To demonstrate the performance of the proposed FMMS for black-box attacking VLP models, we conduct a comprehensive evaluation across significantly different models.
We generate adversarial examples on both two types of VLP models, \emph{i.e.}, fused and aligned VLP models.
The black-box attack performance is evaluated when all other models remain as the target model. Following SGA~\cite{SGA}, we ensure the consistency of image input size.

Table~\ref{Table2} shows the 
comparison with FMMS implemented Top-$N$ search strategy across various VLP models on Flickr30K for TR and IR subtasks. 
The experimental results demonstrate that our proposed FMMS outperforms the existing multimodal attack methods in both white-box and black-box settings. 
Specifically, FMMS achieves a 100\% attack success rate for TR R@1 and outperforms the baselines for IR R@1 across various VLP models in the white-box setting. 
In the black-box setting, our FMMS achieves significant improvements in ASR when the surrogate and target models are of the same type.
For instance, the ASR of adversarial examples generated on ALBEF to attack TCL surpasses Co-attack by approximately 50\% and SGA by about 20\%. 
Moreover, when the surrogate and target models are not of the same type, FMMS still achieves a significant performance improvement. 
For example, when ALBEF is used as the surrogate model to attack CLIP models, it improves by about 30\% compared to Co-attack, while the typical transfer-based attack method SGA improves by only about 10\%. 
Following SGA, the strategy of word replacement used in updating adversarial text introduces significant discreteness in optimizing adversarial text, which results in less improvement in performance on the IR subtask compared to the TR subtask. 
The experimental results on the MSCOCO dataset are similar, as shown in Table ~\ref{Table3}.

\begin{table*}[h]
\small
\setlength{\tabcolsep}{2mm}{
\begin{tabular}{
>{\columncolor[HTML]{FFFFFF}}c 
>{\columncolor[HTML]{FFFFFF}}l 
>{\columncolor[HTML]{FFFFFF}}c 
>{\columncolor[HTML]{FFFFFF}}l 
>{\columncolor[HTML]{FFFFFF}}c 
>{\columncolor[HTML]{FFFFFF}}c 
>{\columncolor[HTML]{FFFFFF}}c 
>{\columncolor[HTML]{FFFFFF}}c 
>{\columncolor[HTML]{FFFFFF}}c 
>{\columncolor[HTML]{FFFFFF}}c 
>{\columncolor[HTML]{FFFFFF}}c 
>{\columncolor[HTML]{FFFFFF}}c }
\hline
\multicolumn{12}{c}{\cellcolor[HTML]{FFFFFF}MSCOCO}                                                                                                                                          \\ \hline
\multicolumn{4}{c|}{\cellcolor[HTML]{FFFFFF}Target Model (\textrightarrow)}                                                                       & \multicolumn{2}{c|}{\cellcolor[HTML]{FFFFFF}ALBEF}                                                                                               & \multicolumn{2}{c|}{\cellcolor[HTML]{FFFFFF}TCL}                                                                           & \multicolumn{2}{c|}{\cellcolor[HTML]{FFFFFF}CLIP$_{\text{ViT}}$}                                                           & \multicolumn{2}{c}{\cellcolor[HTML]{FFFFFF}CLIP$_{\text{CNN}}$}                \\ \hline
\multicolumn{2}{c|}{\cellcolor[HTML]{FFFFFF}Surrogate Model (\textdownarrow)}                        & \multicolumn{2}{c|}{\cellcolor[HTML]{FFFFFF}Attack Method (\textdownarrow)} & TR R@1                                                      & \multicolumn{1}{c|}{\cellcolor[HTML]{FFFFFF}IR R@1}                                & TR R@1                                & \multicolumn{1}{c|}{\cellcolor[HTML]{FFFFFF}IR R@1}                                & TR R@1                                & \multicolumn{1}{c|}{\cellcolor[HTML]{FFFFFF}IR R@1}                                & TR R@1                                & IR R@1                                 \\ \hline
\multicolumn{2}{c|}{\cellcolor[HTML]{FFFFFF}}                                      & \multicolumn{2}{c|}{\cellcolor[HTML]{FFFFFF}PGD}             & \cellcolor[HTML]{EFEFEF}67.54                               & \multicolumn{1}{c|}{\cellcolor[HTML]{EFEFEF}73.72}                                 & 11.08                                 & \multicolumn{1}{c|}{\cellcolor[HTML]{FFFFFF}13.66}                                 & 14.50                                 & \multicolumn{1}{c|}{\cellcolor[HTML]{FFFFFF}22.86}                                 & 17.12                                 & 23.75                                  \\
\multicolumn{2}{c|}{\cellcolor[HTML]{FFFFFF}}                                      & \multicolumn{2}{c|}{\cellcolor[HTML]{FFFFFF}BERT-Attack}     & \cellcolor[HTML]{EFEFEF}35.91                               & \multicolumn{1}{c|}{\cellcolor[HTML]{EFEFEF}47.33}                                 & 35.93                                 & \multicolumn{1}{c|}{\cellcolor[HTML]{FFFFFF}44.85}                                 & 53.22                                 & \multicolumn{1}{c|}{\cellcolor[HTML]{FFFFFF}63.41}                                 & 56.11                                 & 65.01                                  \\
\multicolumn{2}{c|}{\cellcolor[HTML]{FFFFFF}}                                      & \multicolumn{2}{c|}{\cellcolor[HTML]{FFFFFF}Sep-Attack}      & \cellcolor[HTML]{EFEFEF}83.89                               & \multicolumn{1}{c|}{\cellcolor[HTML]{EFEFEF}87.50}                                 & 42.12                                 & \multicolumn{1}{c|}{\cellcolor[HTML]{FFFFFF}51.29}                                 & 54.25                                 & \multicolumn{1}{c|}{\cellcolor[HTML]{FFFFFF}63.88}                                 & 57.13                                 & 65.07                                  \\
\multicolumn{2}{c|}{\cellcolor[HTML]{FFFFFF}}                                      & \multicolumn{2}{c|}{\cellcolor[HTML]{FFFFFF}Co-Attack}       & \cellcolor[HTML]{EFEFEF}79.87                               & \multicolumn{1}{c|}{\cellcolor[HTML]{EFEFEF}87.83}                                 & 32.62                                 & \multicolumn{1}{c|}{\cellcolor[HTML]{FFFFFF}43.09}                                 & 44.89                                 & \multicolumn{1}{c|}{\cellcolor[HTML]{FFFFFF}54.75}                                 & 47.30                                 & 55.64                                  \\
\multicolumn{2}{c|}{\cellcolor[HTML]{FFFFFF}}                                      & \multicolumn{2}{c|}{\cellcolor[HTML]{FFFFFF}SGA}             & \cellcolor[HTML]{EFEFEF}96.80                               & \multicolumn{1}{c|}{\cellcolor[HTML]{EFEFEF}97.20}                                 & 58.86                                 & \multicolumn{1}{c|}{\cellcolor[HTML]{FFFFFF}65.47}                                 & 57.92                                 & \multicolumn{1}{c|}{\cellcolor[HTML]{FFFFFF}65.03}                                 & 58.48                                 & 66.67                                  \\
\multicolumn{2}{c|}{\multirow{-6}{*}{\cellcolor[HTML]{FFFFFF}ALBEF}}               & \multicolumn{2}{c|}{\cellcolor[HTML]{FFFFFF}FMMS (ours)}     & \cellcolor[HTML]{EFEFEF}{\color[HTML]{333333} \textbf{100}} & \multicolumn{1}{c|}{\cellcolor[HTML]{EFEFEF}{\color[HTML]{333333} \textbf{97.92}}} & {\color[HTML]{333333} \textbf{81.90}} & \multicolumn{1}{c|}{\cellcolor[HTML]{FFFFFF}{\color[HTML]{333333} \textbf{69.81}}} & {\color[HTML]{333333} \textbf{74.13}} & \multicolumn{1}{c|}{\cellcolor[HTML]{FFFFFF}{\color[HTML]{333333} \textbf{66.56}}} & {\color[HTML]{333333} \textbf{75.56}} & {\color[HTML]{333333} \textbf{69.70}}  \\ \hline
\multicolumn{2}{c|}{\cellcolor[HTML]{FFFFFF}}                                      & \multicolumn{2}{c|}{\cellcolor[HTML]{FFFFFF}PGD}             & 14.20                                                       & \multicolumn{1}{c|}{\cellcolor[HTML]{FFFFFF}18.41}                                 & \cellcolor[HTML]{EFEFEF}83.33         & \multicolumn{1}{c|}{\cellcolor[HTML]{EFEFEF}85.61}                                 & 14.46                                 & \multicolumn{1}{c|}{\cellcolor[HTML]{FFFFFF}23.05}                                 & 17.12                                 & 23.51                                  \\
\multicolumn{2}{c|}{\cellcolor[HTML]{FFFFFF}}                                      & \multicolumn{2}{c|}{\cellcolor[HTML]{FFFFFF}BERT-Attack}     & 36.04                                                       & \multicolumn{1}{c|}{\cellcolor[HTML]{FFFFFF}47.36}                                 & \cellcolor[HTML]{EFEFEF}37.67         & \multicolumn{1}{c|}{\cellcolor[HTML]{EFEFEF}48.00}                                 & 55.28                                 & \multicolumn{1}{c|}{\cellcolor[HTML]{FFFFFF}63.76}                                 & 57.13                                 & 65.92                                  \\
\multicolumn{2}{c|}{\cellcolor[HTML]{FFFFFF}}                                      & \multicolumn{2}{c|}{\cellcolor[HTML]{FFFFFF}Sep-Attack}      & 44.83                                                       & \multicolumn{1}{c|}{\cellcolor[HTML]{FFFFFF}56.18}                                 & \cellcolor[HTML]{EFEFEF}91.85         & \multicolumn{1}{c|}{\cellcolor[HTML]{EFEFEF}92.75}                                 & 54.71                                 & \multicolumn{1}{c|}{\cellcolor[HTML]{FFFFFF}64.72}                                 & 58.11                                 & 65.93                                  \\
\multicolumn{2}{c|}{\cellcolor[HTML]{FFFFFF}}                                      & \multicolumn{2}{c|}{\cellcolor[HTML]{FFFFFF}Co-Attack}       & 46.08                                                       & \multicolumn{1}{c|}{\cellcolor[HTML]{FFFFFF}57.09}                                 & \cellcolor[HTML]{EFEFEF}85.38         & \multicolumn{1}{c|}{\cellcolor[HTML]{EFEFEF}91.39}                                 & 51.62                                 & \multicolumn{1}{c|}{\cellcolor[HTML]{FFFFFF}60.46}                                 & 52.13                                 & 62.49                                  \\
\multicolumn{2}{c|}{\cellcolor[HTML]{FFFFFF}}                                      & \multicolumn{2}{c|}{\cellcolor[HTML]{FFFFFF}SGA}             & 65.25                                                       & \multicolumn{1}{c|}{\cellcolor[HTML]{FFFFFF}73.20}                                 & \cellcolor[HTML]{EFEFEF}99.07         & \multicolumn{1}{c|}{\cellcolor[HTML]{EFEFEF}99.13}                                 & 56.16                                 & \multicolumn{1}{c|}{\cellcolor[HTML]{FFFFFF}63.70}                                 & 59.22                                 & 65.58                                  \\
\multicolumn{2}{c|}{\multirow{-6}{*}{\cellcolor[HTML]{FFFFFF}TCL}}                 & \multicolumn{2}{c|}{\cellcolor[HTML]{FFFFFF}FMMS (ours)}     & \textbf{82.94}                                              & \multicolumn{1}{c|}{\cellcolor[HTML]{FFFFFF}\textbf{74.93}}                        & \cellcolor[HTML]{EFEFEF}\textbf{100}  & \multicolumn{1}{c|}{\cellcolor[HTML]{EFEFEF}\textbf{99.40}}                        & \textbf{74.28}                        & \multicolumn{1}{c|}{\cellcolor[HTML]{FFFFFF}\textbf{66.50}}                        & \textbf{76.91}                        & \textbf{68.66}                         \\ \hline
\multicolumn{2}{l|}{\cellcolor[HTML]{FFFFFF}}                                      & \multicolumn{2}{c|}{\cellcolor[HTML]{FFFFFF}PGD}             & 7.12                                                        & \multicolumn{1}{c|}{\cellcolor[HTML]{FFFFFF}10.48}                                 & 8.20                                  & \multicolumn{1}{c|}{\cellcolor[HTML]{FFFFFF}11.78}                                 & \cellcolor[HTML]{EFEFEF}54.64         & \multicolumn{1}{c|}{\cellcolor[HTML]{EFEFEF}67.02}                                 & 7.85                                  & 11.25                                  \\
\multicolumn{2}{l|}{\cellcolor[HTML]{FFFFFF}}                                      & \multicolumn{2}{c|}{\cellcolor[HTML]{FFFFFF}BERT-Attack}     & 23.38                                                       & \multicolumn{1}{c|}{\cellcolor[HTML]{FFFFFF}34.39}                                 & 24.66                                 & \multicolumn{1}{c|}{\cellcolor[HTML]{FFFFFF}34.21}                                 & \cellcolor[HTML]{EFEFEF}42.88         & \multicolumn{1}{c|}{\cellcolor[HTML]{EFEFEF}47.14}                                 & 42.38                                 & 48.49                                  \\
\multicolumn{2}{l|}{\cellcolor[HTML]{FFFFFF}}                                      & \multicolumn{2}{c|}{\cellcolor[HTML]{FFFFFF}Sep-Attack}      & 23.00                                                       & \multicolumn{1}{c|}{\cellcolor[HTML]{FFFFFF}34.30}                                 & 25.05                                 & \multicolumn{1}{c|}{\cellcolor[HTML]{FFFFFF}34.80}                                 & \cellcolor[HTML]{EFEFEF}69.10         & \multicolumn{1}{c|}{\cellcolor[HTML]{EFEFEF}78.64}                                 & 42.09                                 & 50.04                                  \\
\multicolumn{2}{c|}{\cellcolor[HTML]{FFFFFF}}                                      & \multicolumn{2}{c|}{\cellcolor[HTML]{FFFFFF}Co-Attack}       & 30.28                                                       & \multicolumn{1}{c|}{\cellcolor[HTML]{FFFFFF}42.67}                                 & 32.84                                 & \multicolumn{1}{c|}{\cellcolor[HTML]{FFFFFF}44.69}                                 & \cellcolor[HTML]{EFEFEF}97.98         & \multicolumn{1}{c|}{\cellcolor[HTML]{EFEFEF}98.80}                                 & 55.08                                 & 62.51                                  \\
\multicolumn{2}{l|}{\cellcolor[HTML]{FFFFFF}}                                      & \multicolumn{2}{c|}{\cellcolor[HTML]{FFFFFF}SGA}             & 33.23                                                       & \multicolumn{1}{c|}{\cellcolor[HTML]{FFFFFF}44.53}                                 & 34.31                                 & \multicolumn{1}{c|}{\cellcolor[HTML]{FFFFFF}45.82}                                 & \cellcolor[HTML]{EFEFEF}99.85         & \multicolumn{1}{c|}{\cellcolor[HTML]{EFEFEF}\textbf{99.85}}                        & 58.44                                 & 66.12                                  \\
\multicolumn{2}{c|}{\multirow{-6}{*}{\cellcolor[HTML]{FFFFFF}CLIP$_{\text{ViT}}$}} & \multicolumn{2}{c|}{\cellcolor[HTML]{FFFFFF}FMMS (ours)}     & \textbf{48.05}                                              & \multicolumn{1}{c|}{\cellcolor[HTML]{FFFFFF}\textbf{46.18}}                        & \textbf{49.50}                        & \multicolumn{1}{c|}{\cellcolor[HTML]{FFFFFF}\textbf{47.34}}                        & \cellcolor[HTML]{EFEFEF}\textbf{100}  & \multicolumn{1}{c|}{\cellcolor[HTML]{EFEFEF}99.84}                                 & \textbf{79.28}                        & \textbf{67.80}                         \\ \hline
\multicolumn{2}{l|}{\cellcolor[HTML]{FFFFFF}}                                      & \multicolumn{2}{c|}{\cellcolor[HTML]{FFFFFF}PGD}             & 6.65                                                        & \multicolumn{1}{c|}{\cellcolor[HTML]{FFFFFF}10.40}                                 & 8.10                                  & \multicolumn{1}{c|}{\cellcolor[HTML]{FFFFFF}11.40}                                 & 5.11                                  & \multicolumn{1}{c|}{\cellcolor[HTML]{FFFFFF}10.05}                                 & \cellcolor[HTML]{EFEFEF}76.95         & \cellcolor[HTML]{EFEFEF}85.22          \\
\multicolumn{2}{l|}{\cellcolor[HTML]{FFFFFF}}                                      & \multicolumn{2}{c|}{\cellcolor[HTML]{FFFFFF}BERT-Attack}     & 26.27                                                       & \multicolumn{1}{c|}{\cellcolor[HTML]{FFFFFF}38.86}                                 & 28.15                                 & \multicolumn{1}{c|}{\cellcolor[HTML]{FFFFFF}39.18}                                 & 50.48                                 & \multicolumn{1}{c|}{\cellcolor[HTML]{FFFFFF}54.57}                                 & \cellcolor[HTML]{EFEFEF}51.61         & \cellcolor[HTML]{EFEFEF}57.71          \\
\multicolumn{2}{l|}{\cellcolor[HTML]{FFFFFF}}                                      & \multicolumn{2}{c|}{\cellcolor[HTML]{FFFFFF}Sep-Attack}      & 26.35                                                       & \multicolumn{1}{c|}{\cellcolor[HTML]{FFFFFF}38.70}                                 & 28.54                                 & \multicolumn{1}{c|}{\cellcolor[HTML]{FFFFFF}39.64}                                 & 50.10                                 & \multicolumn{1}{c|}{\cellcolor[HTML]{FFFFFF}56.52}                                 & \cellcolor[HTML]{EFEFEF}89.66         & \cellcolor[HTML]{EFEFEF}93.00          \\
\multicolumn{2}{l|}{\cellcolor[HTML]{FFFFFF}}                                      & \multicolumn{2}{c|}{\cellcolor[HTML]{FFFFFF}Co-Attack}       & 29.83                                                       & \multicolumn{1}{c|}{\cellcolor[HTML]{FFFFFF}41.97}                                 & 32.97                                 & \multicolumn{1}{c|}{\cellcolor[HTML]{FFFFFF}43.72}                                 & 53.10                                 & \multicolumn{1}{c|}{\cellcolor[HTML]{FFFFFF}58.90}                                 & \cellcolor[HTML]{EFEFEF}96.72         & \cellcolor[HTML]{EFEFEF}98.56          \\
\multicolumn{2}{l|}{\cellcolor[HTML]{FFFFFF}}                                      & \multicolumn{2}{c|}{\cellcolor[HTML]{FFFFFF}SGA}             & 31.61                                                       & \multicolumn{1}{c|}{\cellcolor[HTML]{FFFFFF}43.03}                                 & 32.72                                 & \multicolumn{1}{c|}{\cellcolor[HTML]{FFFFFF}44.28}                                 & 57.00                                 & \multicolumn{1}{c|}{\cellcolor[HTML]{FFFFFF}60.35}                                 & \cellcolor[HTML]{EFEFEF}99.51         & \cellcolor[HTML]{EFEFEF}\textbf{99.80} \\
\multicolumn{2}{c|}{\multirow{-6}{*}{\cellcolor[HTML]{FFFFFF}CLIP$_{\text{CNN}}$}} & \multicolumn{2}{c|}{\cellcolor[HTML]{FFFFFF}FMMS (ours)}     & \textbf{43.90}                                              & \multicolumn{1}{c|}{\cellcolor[HTML]{FFFFFF}\textbf{44.44}}                        & \textbf{49.29}                        & \multicolumn{1}{c|}{\cellcolor[HTML]{FFFFFF}\textbf{47.38}}                        & \textbf{72.11}                        & \multicolumn{1}{c|}{\cellcolor[HTML]{FFFFFF}\textbf{62.04}}                        & \cellcolor[HTML]{EFEFEF}\textbf{100}  & \cellcolor[HTML]{EFEFEF}99.79          \\ \hline
\end{tabular}%
}

\caption{Comparison with existing attack methods on the image-text retrieval task on MSCOCO. It presents the attack success rate (ASR, \%) R@1 for TR and IR. The gray background cells in the table indicate white-box attacks. A higher ASR in black-box attacks denotes better adversarial performance. The best results appear in \textbf{bold}.}
\label{Table3}
\end{table*}
\begin{table*}[!htbp]
\small
\begin{tabular}{
>{\columncolor[HTML]{FFFFFF}}c 
>{\columncolor[HTML]{FFFFFF}}l 
>{\columncolor[HTML]{FFFFFF}}c 
>{\columncolor[HTML]{FFFFFF}}l |
>{\columncolor[HTML]{FFFFFF}}c 
>{\columncolor[HTML]{FFFFFF}}c |
>{\columncolor[HTML]{FFFFFF}}c 
>{\columncolor[HTML]{FFFFFF}}c |
>{\columncolor[HTML]{FFFFFF}}c 
>{\columncolor[HTML]{FFFFFF}}c |
>{\columncolor[HTML]{FFFFFF}}c 
>{\columncolor[HTML]{FFFFFF}}c }
\hline
\multicolumn{4}{c|}{\cellcolor[HTML]{FFFFFF}Target Model (\textrightarrow)}                                                                                             & \multicolumn{2}{c|}{\cellcolor[HTML]{FFFFFF}ALBEF}          & \multicolumn{2}{c|}{\cellcolor[HTML]{FFFFFF}TCL}            & \multicolumn{2}{c|}{\cellcolor[HTML]{FFFFFF}CLIP$_{\text{ViT}}$} & \multicolumn{2}{c}{\cellcolor[HTML]{FFFFFF}CLIP$_{\text{CNN}}$} \\ \hline
\multicolumn{2}{c|}{\cellcolor[HTML]{FFFFFF}Surrogate Model (\textdownarrow)} & \multicolumn{2}{c|}{\cellcolor[HTML]{FFFFFF}FMMS Type (\textdownarrow)} & TR R@1                      & IR R@1                        & TR R@1                      & IR R@1                        & TR R@1                         & IR R@1                          & TR R@1                        & IR R@1                          \\ \hline
\multicolumn{2}{c|}{\cellcolor[HTML]{FFFFFF}}                                             & \multicolumn{2}{c|}{\cellcolor[HTML]{FFFFFF}Full}                                   & \cellcolor[HTML]{EFEFEF}100 & \cellcolor[HTML]{EFEFEF}97.85 & 65.23                       & 57.48                         & 50.06                          & 45.10                           & 53.31                         & 49.09                           \\
\multicolumn{2}{c|}{\multirow{-2}{*}{\cellcolor[HTML]{FFFFFF}ALBEF}}                      & \multicolumn{2}{c|}{\cellcolor[HTML]{FFFFFF}Top-N}                                    & \cellcolor[HTML]{EFEFEF}100 & \cellcolor[HTML]{EFEFEF}97.85 & \textbf{67.97}              & \textbf{58.88}                & \textbf{50.43}                 & \textbf{46.59}                  & \textbf{53.38}                & \textbf{50.57}                  \\ \hline
\multicolumn{2}{c|}{\cellcolor[HTML]{FFFFFF}}                                             & \multicolumn{2}{c|}{\cellcolor[HTML]{FFFFFF}Full}                                   & 64.86                       & 60.17                         & \cellcolor[HTML]{EFEFEF}100 & \cellcolor[HTML]{EFEFEF}99.33 & 48.71                          & 45.84                           & 54.02                         & 49.47                           \\
\multicolumn{2}{c|}{\multirow{-2}{*}{\cellcolor[HTML]{FFFFFF}TCL}}                        & \multicolumn{2}{c|}{\cellcolor[HTML]{FFFFFF}Top-N}                                    & \textbf{69.43}              & \textbf{61.48}                & \cellcolor[HTML]{EFEFEF}100 & \cellcolor[HTML]{EFEFEF}99.36 & \textbf{50.67}                 & \textbf{47.68}                  & \textbf{57.34}                & \textbf{51.01}                  \\ \hline
\multicolumn{2}{c|}{\cellcolor[HTML]{FFFFFF}}                                             & \multicolumn{2}{c|}{\cellcolor[HTML]{FFFFFF}Full}                                   & 18.46                       & 26.64                         & 20.76                       & \textbf{29.40}                & \cellcolor[HTML]{EFEFEF}100    & \cellcolor[HTML]{EFEFEF}99.36   & 57.09                         & 46.72                           \\
\multicolumn{2}{c|}{\multirow{-2}{*}{\cellcolor[HTML]{FFFFFF}CLIP$_{\text{ViT}}$}}        & \multicolumn{2}{c|}{\cellcolor[HTML]{FFFFFF}Top-N}                                    & \textbf{20.75}              & \textbf{27.27}                & \textbf{22.55}              & 29.14                         & \cellcolor[HTML]{EFEFEF}100    & \cellcolor[HTML]{EFEFEF}99.07   & \textbf{59.00}                & \textbf{49.13}                  \\ \hline
\multicolumn{2}{c|}{\cellcolor[HTML]{FFFFFF}}                                             & \multicolumn{2}{c|}{\cellcolor[HTML]{FFFFFF}Full}                                   & 16.37                       & 25.09                         & \textbf{20.02}              & \textbf{28.45}                & 45.40                          & 42.48                           & \cellcolor[HTML]{EFEFEF}100   & \cellcolor[HTML]{EFEFEF}99.59   \\
\multicolumn{2}{c|}{\multirow{-2}{*}{\cellcolor[HTML]{FFFFFF}CLIP$_{\text{CNN}}$}}        & \multicolumn{2}{c|}{\cellcolor[HTML]{FFFFFF}Top-N}                                    & \textbf{17.73}              & \textbf{25.77}                & 19.60                       & 27.88                         & \textbf{47.48}                 & \textbf{44.10}                  & \cellcolor[HTML]{EFEFEF}100   & \cellcolor[HTML]{EFEFEF}99.36   \\ \hline
\end{tabular}%

\caption{Comparison on two FMMS strategies on the image-text retrieval task on Flickr30K. It presents the attack success rate (ASR, \%) R@1 for TR and IR. The gray background cells in the table indicate white-box attacks. A higher ASR in black-box attacks denotes better adversarial performance. The best results appear in \textbf{bold}.}
\label{Table4}
\end{table*}
\subsubsection{Comparison on the Full and Top-$N$ Strategies}
 
 Furthermore, we explore the attack performance of FMMS under the Full search strategy and the Top-$N$ search strategy. Table~\ref{Table4} shows the attack success rate (ASR) of FMMS across various VLP models on Flickr30K under different search strategies. It can be observed that the ASR of the Top-$N$ search strategy is almost higher than that of the Full strategy in black-box settings. 
 It indicates that the Top-$N$ search strategy leverages more feedback information to effectively narrow the search space compared to the Full search, achieving a better attack success rate (ASR). Specifically, when ALBEF, TCL, and CLIP$_{\text{ViT}}$ are used as surrogate models to attack various VLP models, the black-box performance of Top-$N$ FMMS consistently achieves the best results. Overall, our proposed Full FMMS and Top-$N$ FMMS significantly outperform the current typical transfer-based attacks, validating the outstanding performance of our approach.

\subsection{Further Discussion}
\subsubsection{Hyperparameter Study} 
\begin{figure}[!htbp]
\centering
\includegraphics[width=3.3in]{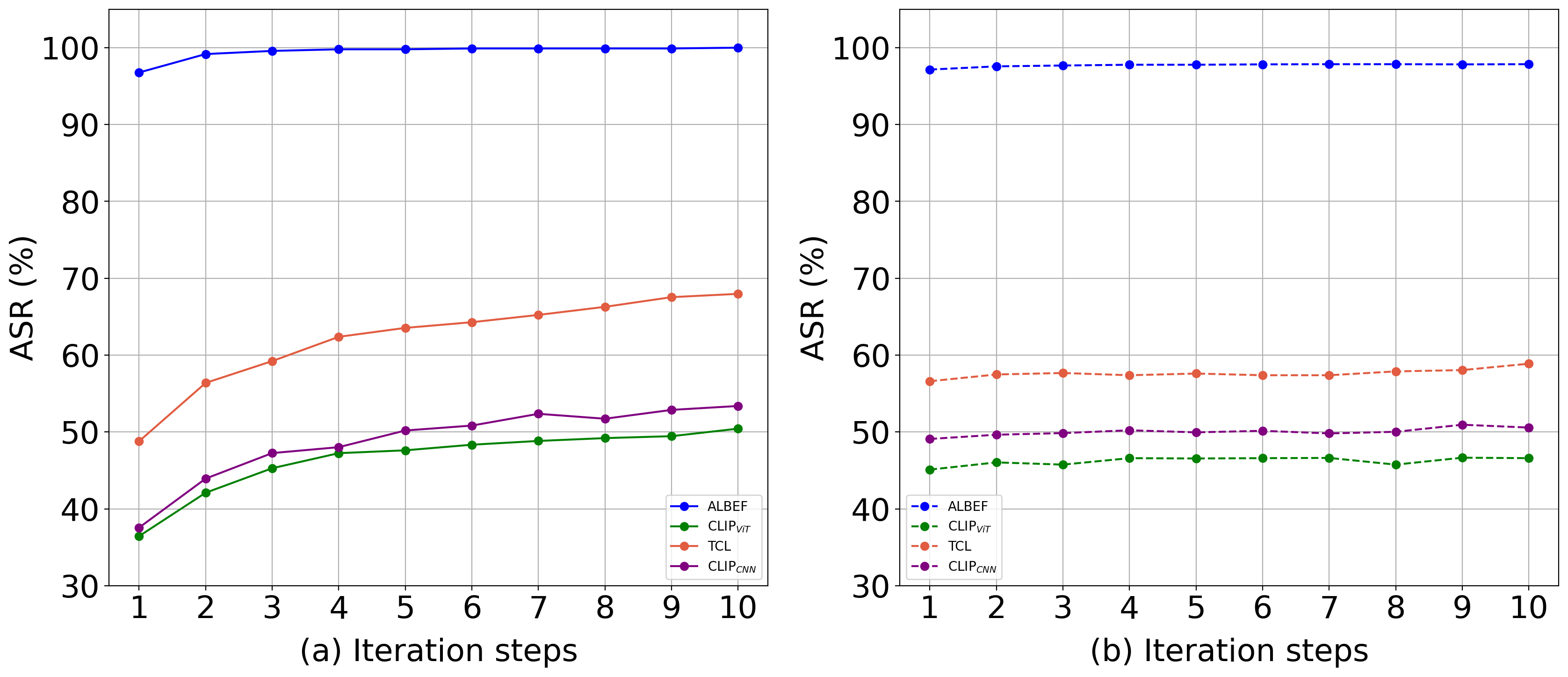}
\caption{
The attack success rate (ASR) with the number of iterations on different target models. Adversarial examples are generated on ALBEF. 
(a) represents the TR R@1 ASR and 
(b) represents the IR R@1 ASR.  
}
\label{discussion}
\end{figure}

We further analyze the relationship between the attack success rate and iteration steps on the two subtasks of image-text retrieval, as shown in Figure~\ref{discussion}. It is reasonable to observe that the attack performance becomes more effective with an increasing number of iterations for both TR and IR tasks. 
When the number of iterations $T>8$, the attack performance of FMMS starts to converge. Therefore, we set the iteration steps $T=10$ to 
trade off the attack success rate and computational costs. 

\subsubsection{Limitation} 
The ASR of FMMS in the TR subtask increases significantly with more iterations, while in the IR subtask, the increase is marginal.
This difference arises because updating adversarial images is a continuous optimization, making ASR highly proportional to iteration count. 
However, the process of optimizing the adversarial text is discrete, with word-level replacements causing instability during the search. Consequently, as depicted 
in Figure~\ref{discussion}, the ASR in the TR subtask shows a marked promotion, whereas in the IR subtask, it improves slightly and oscillates. 


\section{Conclusion}
The feature distributions across different vision-language pre-training (VLP) models can vary significantly, making it hard to generate well-transferable adversarial examples. To this end, we propose a novel attack paradigm, namely Feedback-based Modal Mutual Search (FMMS), to attack VLP models.
We introduce a modal mutual loss for cross-modal interactions, guiding the gradient of the adversarial examples with a more promising direction.
By leveraging the target model's feedback for multi-round refinement, FMMS further optimizes the adversarial examples, driving unsuccessful examples into the adversarial region.
Based on the VLP model’s feedback information, 
we present two variants of FMMS, \textit{i.e.,} Full FMMS and Top-$N$ FMMS,
where the Top-$N$ version narrows the search space compared to the Full version.
Extensive experimental results demonstrate that FMMS outperforms existing baselines. 


\clearpage
\bibliography{aaai25}
\end{document}